\newif\ifRAL
\RALfalse 	

\newif\ifTR
\TRfalse 	

\newif\ifPrePrint
\PrePrintfalse

\newif\ifDraft
\Draftfalse

\ifRAL
\documentclass[letterpaper, 10 pt, journal, twoside]{IEEEtran}
\else
\documentclass[letterpaper, 10 pt, conference]{ieeeconf}
\fi

\IEEEoverridecommandlockouts

\ifRAL
\else
\fi		

\makeatletter

\let\proof\@undefined
\let\endproof\@undefined
\makeatother

\usepackage{amsmath} 
\usepackage{amssymb}  
\usepackage{amsthm}
\usepackage{amsfonts}
\usepackage{mathtools}
\usepackage{acro}
\usepackage{bm}
\usepackage{cite}
\providecommand{\bm}{\pmb}

\theoremstyle{definition}

\theoremstyle{remark}

\usepackage{verbatim}

\usepackage{color}
\usepackage{subcaption}
\usepackage{graphicx}
\usepackage{caption}
\usepackage{siunitx}

\usepackage{optidef}

\usepackage{times}

\usepackage{xspace}

\usepackage[ruled,vlined,linesnumbered]{algorithm2e}
\usepackage[noend]{algorithmic}

\graphicspath{{images/}}

\usepackage[us]{datetime}
\usepackage{caption}
\usepackage[usenames,dvipsnames,table]{xcolor}

\usepackage{hyperref} 
\hypersetup{
    colorlinks,
    citecolor=black,
    filecolor=black,
    linkcolor=black,
    urlcolor=black,
    pdfauthor={},
    pdfsubject={},
    pdftitle={}
}
\usepackage[capitalise]{cleveref}
\crefformat{equation}{(#2#1#3)}
\let\labelindent\relax
\usepackage[inline]{enumitem}

\usepackage{todonotes}

\usepackage{graphicx}

\usepackage{latexsym}
\usepackage{color}

\usepackage{cite}

\usepackage{caption}

\usepackage{tikz}
\usetikzlibrary{shapes,arrows}
\usetikzlibrary{arrows.meta}
\usetikzlibrary{positioning,fit,backgrounds}

\tikzstyle{medium box}=[fill=white, draw=black, shape=rectangle, minimum width=4cm, minimum height=0.75cm]
\tikzstyle{bigBox}=[fill=white, draw=black, shape=rectangle, minimum width=1.2cm, minimum height=0.8cm]
\tikzstyle{smallBox}=[fill=white, draw=black, shape=rectangle, minimum width=1.0cm, minimum height=0.875cm]
\tikzstyle{smallmediumbox}=[fill=white, draw=black, shape=rectangle, minimum width=3cm, minimum height=0.75cm]
\tikzstyle{onelinebox}=[fill=white, draw=black, shape=rectangle, minimum width=1cm, minimum height=0.5cm]
\tikzstyle{arrow edge}=[-Latex]
\tikzstyle{none}=[inner sep=0mm]
\tikzstyle{whitebox}=[fill=white, draw=white, shape=rectangle]
\tikzstyle{circle}=[fill=white, draw=black, shape=circle, minimum size=6pt, inner sep=0pt]

\tikzstyle{arrow}=[->]

\usepackage{tabularx, booktabs}
\newcolumntype{Y}{>{\centering\arraybackslash}X}
\usepackage{multirow}

\usepackage{color}

\renewcommand{\vec}[1]{\bm{#1}}		
\newcommand{\mat}[1]{\bm{#1}}		
\DeclareMathOperator*{\argmin}{arg\,min}
\newcommand{\nR}[1]{\mathbb{R}^{#1}}		
\renewcommand{\matrix}[1]{\begin{bmatrix} #1 \end{bmatrix}}	
\newcommand{\upperRomannumeral}[1]{\uppercase\expandafter{\romannumeral#1}}	

\newcommand{\transpose}{^\top}
\newcommand{\des}{r}
\newcommand{\desk}{r,k}
\newcommand{\ddt}{\frac{d}{dt}}
\newcommand{\SO}{\mathrm{SO}}
\newcommand{\f}[1]{\prescript{}{#1}{}}

\renewcommand{\frame}[1]{\mathcal{F}_{#1}}		
\newcommand{\pos}{\vec{p}}				
\newcommand{\vel}{\vec{v}}				
\newcommand{\rotMat}{\mat{R}}				
\newcommand{\eye}[1]{\mat{I}_{#1}}
\newcommand{\zeros}[1]{\mat{0}_{#1}}

\newcommand{\body}{{}_B}
\newcommand{\local}{{}_L}
\newcommand{\world}{{}_W}

\newcommand{\origin}{O}
\newcommand{\vX}{\vec{x}}					
\newcommand{\vY}{\vec{y}}					
\newcommand{\vZ}{\vec{z}}					

\newcommand{\meas}{{}_m}

\newcommand{\originW}{\origin\world}
\newcommand{\xW}{\vX\world}				
\newcommand{\yW}{\vY\world}				
\newcommand{\zW}{\vZ\world}				

\newcommand{\originB}{\origin\body}
\newcommand{\xB}{\vX\body}				
\newcommand{\yB}{\vY\body}				
\newcommand{\zB}{\vZ\body}				

\newcommand{\originL}{\origin\local}
\newcommand{\xL}{\vX\local}				
\newcommand{\yL}{\vY\local}				
\newcommand{\zL}{\vZ\local}				

\newcommand{\RB}{\mat{R}_{B}}

\newcommand{\states}{\vec{x}}
\newcommand{\inputs}{\vec{u}}
\newcommand{\stateSpace}{\mathcal{X}}
\newcommand{\inputSpace}{\mathcal{U}}

\newcommand{\angAcc}{\dot{\angVel}}		
\newcommand{\gravity}{\vec{g}}			
\newcommand{\tprop}{t}
\newcommand{\tprops}{\vec{t}}
\newcommand{\tiltangles}{\vec{\alpha}}
\newcommand{\tpropsdot}{\dot{\vec{t}}}
\newcommand{\tiltanglesdot}{\dot{\vec{\alpha}}}

\newcommand{\qB}{\vec{q}}   

\newcommand{\angVel}{\vec{\omega}}		

\newcommand{\mass}{m}
\newcommand{\inertia}{\bm{J}}

\newcommand{\force}{\vec{f}}
\newcommand{\torque}{\vec{\tau}}
\newcommand{\actForce}{\vec{f}_a}
\newcommand{\actTorque}{\vec{\tau}_a}
\newcommand{\actWrench}{\vec{w}_a}
\newcommand{\actWrenchDot}{\dot{\vec{w}}_a}
\newcommand{\actInput}{\inputs_a}
\newcommand{\actInputDot}{\dot{\inputs}_a}
\newcommand{\wrench}{\vec{w}}

\newcommand{\imuLinAccMeas}{\vec{a}_{IMU}}

\newcommand{\imuAngAccMeas}{\dot{\vec{\omega}}_{IMU}}
\newcommand{\imuAngVel}{\vec{\omega}_{IMU}}

\newcommand{\imuAngAcc}{\dot{\vec{\omega}}_{IMU}}

\newcommand{\features}{\tilde{\vec{x}}}
\newcommand{\modelMatrix}{\mat{C}}

\newcommand{\tiltAngles}{\vec{\alpha}}
\newcommand{\dtiltAngles}{\dot{\vec{\alpha}}}

\newcommand{\drotorSpeeds}{\dot{\vec{\omega}}}
\newcommand{\thrustComponents}{\tilde{\vec{t}}}

\newcommand{\nrot}{n_r}
\newcommand{\nrotperarm}{n_{rpa}}
\newcommand{\narms}{n_a}

\newcommand{\allocationMatrix}{\mat{A}}

\newcommand{\qErr}{\vec{q}_{\text{e}}}
\newcommand{\qErrk}{\vec{q}_{\text{e},k}}
\newcommand{\RBdes}{\mat{R}_{B,\des}}
\newcommand{\discreteDynamics}{\vec{g}}
\newcommand{\stageCost}{\vec{h}}

\newcommand{\nstates}{n}
\newcommand{\ninputs}{m}
\newcommand{\nfeatures}{n_f}
\newcommand{\nsamples}{n_s}

\newcommand{\stateEst}{\hat{\vec{x}}}
\newcommand{\posEst}{\hat{\pos}}
\newcommand{\velEst}{\hat{\vel}}
\newcommand{\qBEst}{\hat{\qB}}
\newcommand{\ekfInputs}{\vec{u}_{EKF}}
\newcommand{\ekfMeasurements}{\vec{z}_{EKF}}
\newcommand{\posMeas}{\pos\meas}
\newcommand{\qBMeas}{\qB\meas}
\newcommand{\distTorqueEst}{\Delta\hat{\torque}}
\newcommand{\distForceEst}{\Delta\hat{\force}_L}
\newcommand{\angVelEst}{\hat{\vec{\omega}}}		
\newcommand{\processNoise}{\vec{n}}

\newcommand{\torqueResidual}{\Delta\torque}
\newcommand{\torqueResidualMeasured}{\Delta\torque\meas}
\newcommand{\forceResidual}{\Delta\force}
\newcommand{\forceResidualEst}{\Delta\hat{\force}}
\newcommand{\forceResidualMeasured}{\Delta\force\meas}

\newcommand{\wrenchResidual}{\Delta\wrench}
\newcommand{\wrenchResidualMeasured}{\Delta\wrench\meas}
\newcommand{\wrenchResidualModel}{\Delta\bar{\wrench}}

\newcommand{\trainError}{\vec{e}_{train}}

\newcommand{\featureMatrix}{\mat{X}}
\newcommand{\regParam}{\lambda}

\newcommand{\weightThrust}{w_T}
\newcommand{\weightAlpha}{w_\alpha}
\newcommand{\weightAlphaDot}{w_{\dot{\alpha}}}
\newcommand{\mpcStateWeights}{\mat{Q}}
\newcommand{\mpcInputWeights}{\mat{R}}
\newcommand{\actForceConstraint}{f_{a,max}}
\newcommand{\actTorqueConstraint}{\tau_{a,max}}
\newcommand{\actWrenchConstraint}{\wrench_{a,max}}
\newcommand{\actWrenchDotConstraint}{\dot{\wrench}_{a,max}}
\newcommand{\tpropsMin}{\vec{t}_{min}}
\newcommand{\tpropsMax}{\vec{t}_{max}}

\DeclareSIUnit{\amperehour}{Ah}

\DeclareAcronym{MAV}{short = MAV, long = micro aerial vehicle}
\DeclareAcronym{AM}{short = AM, long = aerial manipulator}
\DeclareAcronym{COM}{short = CoM, long = center of mass}
\DeclareAcronym{DOF}{short = DoF, long = degrees of freedom}
\DeclareAcronym{NDT}{short = NDT, long = nondestructive testing}
\DeclareAcronym{OMAV}{short = OMAV, long = omnidirectional micro aerial vehicle}
\DeclareAcronym{CAD}{short = CAD, long = computer-aided design}
\DeclareAcronym{IMU}{short = IMU, long = inertial measurement unit}
\DeclareAcronym{TOF}{short = TOF, long = time of flight}
\DeclareAcronym{EKF}{short = EKF, long = extended Kalman filter}
\DeclareAcronym{LQRI}{short = LQRI, long = linear-quadratic regulator with integral action}
\DeclareAcronym{MPC}{short = MPC, long = model predictive control}
\DeclareAcronym{NMPC}{short = NMPC, long = nonlinear model predictive control}
\DeclareAcronym{PID}{short = PID, long = proportional-integral-derivative}
\DeclareAcronym{EE}{short = EE, long = end-effector}
\DeclareAcronym{ROS}{short = ROS, long = Robot Operating System}
\DeclareAcronym{ACADO}{short = ACADO, long = Automatic Control and Dynamic Optimization}
\DeclareAcronym{FT}{short = FT, long = Force-Torque}
\DeclareAcronym{WMPC}{short = WMPC, long = Wrench-MPC}
\DeclareAcronym{AMPC}{short = AMPC, long = Actuator-MPC}
\DeclareAcronym{LBMPC}{short = LBMPC, long = learning-based MPC}

\DeclarePairedDelimiter{\norm}{\lVert}{\rVert} 

\DeclareMathOperator{\atantwo}{atan2}

\ifRAL 
\fi

\author{Maximilian Brunner, Weixuan Zhang, Ahmad Roumie, Marco Tognon, Roland Siegwart
	\ifRAL
		\thanks{Manuscript received: DD,\,MM,\,YY; Revised DD,\,MM,\,YY ; Accepted DD,\,MM,\,YY.}
		\thanks{This paper was recommended for publication by Editor NAME SURNAME upon evaluation of the Associate Editor and Reviewers' comments. } 
	\fi
	\thanks{All authors are with the Autonomous Systems Lab (ASL), ETH Zurich. Corresponding author: {\tt \footnotesize \href{mailto:maximilian.brunner@mavt.ethz.ch}{maximilian.brunner@mavt.ethz.ch}.}}%
	\thanks{This research was partially supported by NCCR Digital Fabrication.}
	\ifRAL
		\thanks{Digital Object Identifier (DOI): see top of this page.}
	\fi
}

\ifRAL 
\title{MPC with Learned Residual Dynamics with Application on Omnidirectional MAVs}

\else 
\title{\bf MPC with Learned Residual Dynamics with Application on Omnidirectional MAVs}
\fi

\ifPrePrint
\makeatletter
\def\ps@titlepagestyle{
	\def\@oddfoot{}\def\@evenfoot{}
	\def\@oddhead{\textcolor{red}{\sf\footnotesize Preprint version, final version at http://ieeexplore.ieee.org/ \hfill IEEE Robotics and Automation Letters 2019}}
	\def\@evenhead{\textcolor{red}{\sf\footnotesize  Preprint version, final version at http://ieeexplore.ieee.org/  \hfill IEEE Robotics and Automation Letters 2019}}%
}%
\makeatother
\makeatletter
\def\ps@headings{
	\def\@oddfoot{\textcolor{red}{\sf\footnotesize  Preprint version, final version at http://ieeexplore.ieee.org/ \hfill \thepage \;\;~\hfill~\hfill IEEE Robotics and Automation Letters 2020}}\def\@evenfoot{\hfill\thepage\hfill}
	\def\@oddhead{}\def\@evenhead{}%
}%
\makeatother
\fi	

\ifDraft
\makeatletter
\def\ps@titlepagestyle{
	\def\@oddfoot{}\def\@evenfoot{}
	\def\@oddhead{\textcolor{red}{\sf Draft version  \hfill Confidential}}
	\def\@evenhead{\textcolor{red}{\sf  Draft version  \hfill Confidential}}%
}%
\makeatother
\makeatletter
\def\ps@headings{
	\def\@oddfoot{\textcolor{red}{\sf  Draft version  \hfill Confidential}}\def\@evenfoot{\hfill\thepage\hfill}
	\def\@oddhead{}\def\@evenhead{}%
}%
\makeatother
\usepackage{draftwatermark}
\SetWatermarkText{Confidential draft}
\SetWatermarkScale{0.6}
\SetWatermarkLightness{0.9} 

\fi	

\begin{document}

\maketitle

\begin{abstract}
The growing field of aerial manipulation often relies on fully actuated or omnidirectional micro aerial vehicles (OMAVs) which can apply arbitrary forces and torques while in contact with the environment. Control methods are usually based on model-free approaches, separating a high-level wrench controller from an actuator allocation. If necessary, disturbances are rejected by online disturbance observers.
However, while being general, this approach often produces sub-optimal control commands and cannot incorporate constraints given by the platform design.
We present two model-based approaches to control OMAVs for the task of trajectory tracking while rejecting disturbances. The first one optimizes wrench commands and compensates model errors by a model learned from experimental data. The second one optimizes low-level actuator commands, allowing to exploit an allocation nullspace and to consider constraints given by the actuator hardware.
The efficacy and real-time feasibility of both approaches is shown and evaluated in real-world experiments.
\end{abstract}


\ifRAL 
\else 
{} 
\fi

\section{INTRODUCTION}\label{sec:intro}

\ifRAL
\IEEEPARstart{T}{he}
\else
The
\fi
advancement of \acp{MAV} in the recent years has come with increasing focus on aerial physical interaction tasks. 
Investigations started with pick-and-place tasks,  continuing with contact-based inspection and push-and-slide operations \cite{Tognon2019,Bodie2019,Bodie2020,Nava2020,Tzoumanikas2020}, all the way to the manipulation of the environment \cite{Lee2020,Lee2021}. 
Tasks and applications involving aerial physical interaction are getting more complex year by year~\cite{9462539}.

Different platforms have been developed to cope with different challenges~\cite{2021-HamUsaSabStaTogFra}, some being generic for research or a broad spectrum of tasks, other being more specialized for certain applications.
Generally, aerial interaction requires a flying platform equipped with a manipulator that is designed to interact according to the desired task. This manipulator can be passive or actively controlled.  Depending on the task, the flying platform needs to meet certain requirements to compensate for wrenches (i.e., forces and torques) which arise during the interaction. While underactuated platforms are capable of compensating for some limited wrenches, \emph{fully actuated} and \emph{overactuated} platforms offer more freedom in this matter.
\emph{Fully actuated} platforms can compensate for any reaction wrenches that appear during interaction while \emph{overactuation} adds the benefit of redundancy. Furthermore, we refer to \emph{Omnidirectional} \acp{MAV} (OMAVs) as platforms that can generate thrust in any direction, providing sufficient lift force to hover in any possible orientation \cite{Hamandi2020}.

However, full actuation and/or overactuation comes with new challenges and opportunities. First, the higher number of actuators gives potential for more unmodeled disturbances, e.g., through inaccurate thrust (or other actuator) mappings or airflow interferences. Second, it increases the complexity of finding optimal control inputs.
Therefore, we intend to explore the possibilities of finding the optimal control inputs for performing a flight task while rejecting internal disturbances.

\begin{figure}[t]
    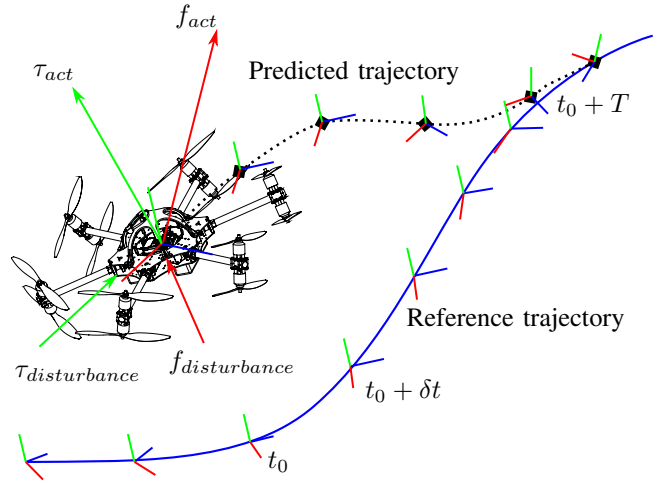
    \caption{Illustration of MPC for trajectory tracking control.}\label{fig:mpc}
\end{figure}

Most common controllers of \acp{MAV} are divided into three parts:
\begin{enumerate*}[label=(\roman*)]
\item a high level controller for pose and/or force tracking (e.g., PD, impedance) which produces linear/angular acceleration commands,\label{item:tracking}
\item a wrench estimator to observe unmodeled disturbances, and\label{item:estimation}
\item an actuator allocation to convert high level commands into actuator commands\label{item:allocation}
\end{enumerate*}.
While this structure has proven to work reliably, it comes with a few drawbacks. First, the separation of a pose tracking controller and the actuator allocation does not allow to fully optimize the actuator commands for the execution of a desired task. Second, employing an online wrench estimator introduces time delays that can impair the actual flight performance.

The controller \ref{item:tracking} initially has often been implemented by a PD or impedance controller that generates acceleration commands based on the tracking errors. More recently, model-based optimal control approaches have been studied as well.
\Ac{MPC} can be specifically useful in situations where accurate and fast trajectory tracking is needed in the presence of actuator constraints and external disturbances~\cite{peric2021direct}. In this context, \cite{Hentzen2019} uses a \ac{MPC} on an underactuated quadrotor in strong wind gusts, comparing different Kalman filters for disturbance estimation. In this work the model based approach leads to higher position tracking accuracy than PID controllers. In a different work, \cite{Torrente2020} models aerodynamic disturbances by Gaussian Processes and fuses them in an \ac{MPC} to improve high-speed flight maneuvers with quadrotors. In \cite{Brunner2020}, we have presented an \ac{MPC}-based control framework for trajectory tracking of \acp{OMAV}. Optimizing on the wrench level, the individual rotor speed and tilt rotor inputs were then found through the same allocation process as presented in \cite{Bodie2019}. 
One difficulty of model-based approaches is their dependence on an accurate model of the system. If this is not known or contains unmodeled disturbances, these errors need to be either estimated online or compensated by an adaptive MPC.

The wrench estimator \ref{item:estimation} can be implemented by a momentum-based observer (MBE) \cite{Ruggiero2014,Bodie2019}, or another estimation framework such as a Kalman filter \cite{Hentzen2019}.
The estimator usually accounts for both \emph{external} as well as \emph{internal} disturbances.  External disturbances can only be modeled to a certain degree (as they are caused by either interaction or unpredictable sources like wind gusts). Internal disturbances on the other hand are caused by unmodeled and unknown effects originating from the system itself. These can result from inaccurate hardware fabrication or complex aerodynamic effects.
A common disadvantage of using an online observer is the inherent time delay as it requires state observations in order to estimate the current disturbance. Therefore, the disturbances can also be learned offline based on experimental data and then applied during a flight. This approach has been applied in \cite{Zhang2020} by using Gaussian Processes to learn the wrench residuals.

Lastly, the actuator allocation \ref{item:allocation} maps a wrench command into the actuator controls that will generate the desired wrench. It is determined by the particular actuation and geometry of the platform. According to the system, the actuator allocation problem can have a unique solution or, if the system is overactuated, an infinite number of solutions. This latter case can be exploited to achieve secondary objectives, like the minimization of energy or the optimization of actuation properties \cite{Ryll2015}.


\subsection{Related works}
In order to cope with the above stated difficulties, various approaches have been investigated.
In \cite{Bicego2020}, the tracking controller and the allocation were in a single optimizer. This framework does not assume any linear model approximations nor does it depend on a cascaded control approach to decouple the translational and rotational dynamics of the rigid body. What is remarkable is the use of the derivatives of the individual propeller forces as control inputs which allows the direct translation of actual control inputs. Nevertheless, in the case of an \acp{OMAV} with actuated tilt angles, there is no explicit relationship between the generated individual forces and the actuator constraints.
\cite{Shawky2021} presented a \ac{NMPC} framework for overactuated MAVs with actively tiltable propellers.  Two optimizers were compared, namely Interior Point Optimization (through IPOPT) and Sequential Quadratic Programming (SQP) in CasADI through ACADO. However, the framework is only evaluated in simulations in Gazebo.

Also different methods to deal with the problem of uncertain or unknown model dynamics have been presented, such as \emph{adaptive MPC}, \emph{robust MPC} (e.g. Tube-MPC \cite{Chisci2001, Langson2004}), or \emph{learning-based} MPC.
A number of approaches exists to improve the MPC model by learning its true dynamics.
\cite{Hewing2020} gives an overview on how MPC performance can be improved through learning from recorded data. Accordingly, this can be achieved by following the following approaches:
\begin{enumerate*}[label=(\roman*)]
\item Learning the system dynamics, and/or\label{item:learning_dynamics}
\item learning the controller design, such as the optimal cost function or constraints\label{item:learning_design}.
\end{enumerate*}
\cite{Piga2019} followed the approach of \emph{Identification for control} (I4C), which aims to not minimize the output prediction errors (i.e. to fit the MPC model to the real system as closely as possible), but rather to find a model that optimizes the control performance in closed loop. In this approach, the MPC acts as an outer loop (i.e., as a \emph{reference governor}) that is based on a model of the inner loop, given by a fast PID controller. In this approach, closed loop experiments are repeated to find optimal control parameters through Bayesian Optimization.
As mentioned in \cite{Lorenzen2019}, ``stochastic and robust MPC are suitable for handling unmodeled dynamics and rapidly changing disturbances''. However, they are conservative and not appropriate for adapting to constant parameters.
In \cite{Aswani2013} the concept of \ac{LBMPC} was introduced and applied on a quadrotor in \cite{Aswani2012}. \Ac{LBMPC} uses a so-called \emph{oracle} to learn the residual dynamics between a model and the true system dynamics. The oracle can be any linear or nonlinear parametric function whose parameters are adapted during the execution of the controller. In \cite{Aswani2012}, an \ac{EKF} is used for joint state and parameter estimation for a linear affine oracle.
Other approaches such as Iterative Learning MPC \cite{Bujarbaruah2018,Tamar2017} are restricted to repetitive tasks, in which the performance can be improved by adapting the control inputs by learning from earlier iterations.

\subsection{Contributions}
In this work, we explore the methods of how to use \ac{MPC} on an \ac{OMAV}. Specifically,  given by its tilt-rotor design, the \ac{OMAV} of our work is capable of a high level of overactuation and potential internal disturbances, leading to model errors.
We present the theory and experimental validation of an actuator-level NMPC that is able to generate real-time rotor speed and tilt angle commands for the task of trajectory tracking. Its knowledge of the actuator allocation allows to explore the actuation nullspace for an overactuated platform while respecting actuator constraints.
Furthermore, we show how internal model disturbances can be learned and applied through a simple linear model inside a wrench-level MPC.
Finally, we compare the performances of the different proposed approaches in various experiments.


\section{Modeling}
In this section we introduce the modeling of the system dynamics using Newton-Euler equations based on the following common assumptions: 
\begin{enumerate*}[label=(\roman*)]
\item The system is a single rigid body,\label{item:rigidbody}
\item mass and inertia matrix are constant,
\item the \ac{COM} coincides with the geometric center of the system, and
\item disturbance forces and moments can be reduced to a single wrench applied to the \ac{COM}.
\end{enumerate*}

\subsection{Notation}
We denote scalars by lowercase symbols, vectors $\vec{v}$ by lowercase bold symbols, and matrices $\mat{M}$ by uppercase bold symbols. If not specified differently, we use subscripts to indicate the frame of a vector $\f{W}\vec{v}$. To represent orientations, we use unit quaternions $\qB=\matrix{q_w & q_x & q_y & q_z}\transpose \in \nR{4}$, such that $\norm{\qB}=1$, as well as rotation matrices $\RB\in \SO(3)$.  Quaternions and rotation matrices can be used interchangeably, such that they act as vector transformations, i.e., $\f{W}\vec{v}=\vec{q}\otimes\f{B}\vec{v}\otimes\vec{q}^{-1}=\RB\f{B}\vec{v}$, where $\otimes$ represents the quaternion multiplication.

\subsection{Frame definitions}
\begin{figure}[ht]
    \centering
    \def\svgwidth{0.9\columnwidth}
    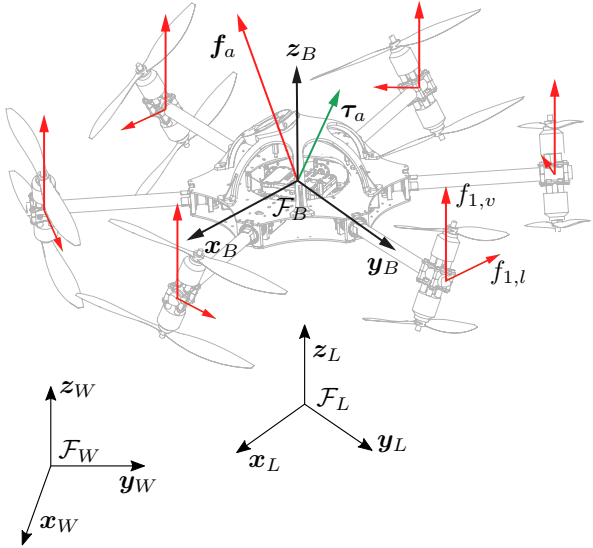\caption{Frame definitions and forces/torques acting on the platform.}\label{fig:frames}
\end{figure}

We will refer to three reference frames: the inertial world frame $\frame{W} = \{\originW, \xW, \yW, \zW  \}$, the body frame $\frame{B} = \{\originB, \xB, \yB, \zB \}$ which is fixed to the geometric center of the \ac{OMAV}, and the local frame $\frame{L}=\{\originL, \xL, \yL, \zL \}$, which is obtained by a pure yaw rotation of the platform yaw angle from the world frame. $\origin_\star$ represents the center of the generic frame $\frame{\star}$, while $(\vX_\star,\vY_\star,\vZ_\star)$ represent its unit axes.
$\frame{W}$ is defined s.t. $\zW$ is aligned with the gravity vector $\gravity=\matrix{0 & 0 &-\SI{9.81}{\meter\per\square\second}}\transpose$.
\Cref{fig:frames} gives an overview of the frames used in this work.


\subsection{Rigid body dynamics}
We define the system states as follows: The position of the \ac{COM} is given by $\pos\in\nR{3}$  in $\frame{W}$ and its velocity by $\vel\in\nR{3}$ in $\frame{B}$.  The attitude is expressed by the unit quaternion $\qB$ and the angular velocity by $\angVel\in\nR{3}$ in $\frame{B}$. The mass and inertia matrix are given by $\mass$ and $\inertia\in\nR{3\times 3}$, respectively.
We assume that we can reduce all forces and torques generated by the actuators to an individual actuator wrench acting on the \ac{COM}, expressed by $\actWrench=\matrix{\actForce\transpose & \actTorque\transpose}\transpose\in\nR{6}$, where $\actForce\in\nR{3}$ and $\actTorque\in\nR{3}$.
%
We can then write the dynamics as follows:
\begin{subequations}
\begin{align} 
    \dot\pos ={}& \RB\vel \\
    \dot{\qB} ={}& \frac{1}{2}\qB\otimes\begin{bmatrix} 0\\ \angVel\end{bmatrix} \\
    \dot\vel ={}& \mass^{-1}(\actForce + \forceResidual) + \RB\transpose\gravity - \angVel\times\vel\\
    \angAcc ={}& \inertia^{-1} \left(\actTorque + \torqueResidual - \angVel \times (\inertia\angVel)\right),\label{eq:angaccbody}%
\end{align}\label{eq:systemDynamics}%
\end{subequations}%
%
where we considered a residual wrench $\wrenchResidual=\matrix{\forceResidual\transpose &\torqueResidual\transpose}\transpose\in\nR{6}$ acting on the \ac{COM}.
The residual (or disturbance) forces and torques $\forceResidual\in\nR{3}$ and $\torqueResidual\in\nR{3}$ account for all unmodeled effects, both internal and external, such as airflow interference (within rotor groups and between different rotor groups), hardware misalignments, or slightly different propeller characteristics. We assume that no further external or time-varying disturbances are present.
Defining a state vector as $\states=\matrix{\pos\transpose & \qB\transpose & \vel\transpose & \angVel\transpose}\transpose$ we can then write the dynamic equations as
\begin{align}
    \dot\states=f_R(\states,\actWrench,\wrenchResidual).\label{eq:systemDynamicsShort}
\end{align}

\subsection{Allocation of actuator commands}

The \ac{OMAV} actuation can be described as follows: There are $\narms$ tiltable arms attached to the body core, with each arm carrying $\nrotperarm$ rotors, resulting in a total number of $\nrot=\narms\nrotperarm$ rotors.
The total actuator wrench $\actWrench$ is the result of the commanded rotor thrusts $\tprops\in\nR{\nrot}$ and the current tilt angle configuration, given by the tilt angles $\tiltAngles\in\nR{\narms}$. The geometry of the platform determines the relation between the actuator commands and the total actuator wrench $\actWrench$.
In the following we present a method for the \emph{actuator allocation} which computes actuator commands from a reference actuator wrench, i.e., $(\tiltAngles,\tprops)=f_{alloc}(\actWrench)$.
\subsubsection*{Actuator allocation}
We define the vector $\thrustComponents(\tiltAngles,\tprops)\in\nR{2\nrot}$ that describes the vertical and lateral thrust components of each propeller in the body frame. We can then write the relation between $\actWrench$ and $\thrustComponents$ by a linear function:
\begin{subequations}
\begin{align}
    \actWrench(\tiltAngles,\tprops)&=\begin{bmatrix}
    \actForce\\
    \actTorque
    \end{bmatrix}=
    \allocationMatrix\thrustComponents(\tiltAngles,\tprops),\label{eq:allocation}\\
    \thrustComponents(\tiltAngles,\tprops)&=\begin{bmatrix}
    f_{1,l}\\
    f_{1,v}\\
    \vdots\\
    f_{\nrot,l}\\
    f_{\nrot,v}
    \end{bmatrix}=\begin{bmatrix}
    \sin(\alpha_1)\tprop_1\\
    \cos(\alpha_1)\tprop_1\\
    \vdots\\
    \sin(\alpha_{\narms})\tprop_{\nrot}\\
    \cos(\alpha_{\narms})\tprop_{\nrot}
    \end{bmatrix}.\label{eq:extendedThrust}
\end{align}
\end{subequations}
Given this relation, the allocation matrix $\allocationMatrix\in\nR{6\times 2\nrot}$ is constant and can be obtained from the platform geometry.
In order to compute the actuator commands $\actInput\coloneqq\matrix{\tiltangles\transpose & \tprops\transpose}\transpose$ from a given actuator wrench $\actWrench$, we apply the Moore-Penrose Inverse:
\begin{subequations}
\begin{align}
\thrustComponents&=\allocationMatrix^\dagger\actWrench+\left(\eye{}-\allocationMatrix^\dagger\allocationMatrix\right)\vec{b}\label{eq:allocationGeneral}\\
\alpha_i&=\atantwo\left(\sum_j^{\nrotperarm} f_{j,l},\sum_j^{\nrotperarm} f_{j,v}\right)\quad\forall i=1\dots \narms\label{eq:alpha}\\
t_i&=\sqrt{f_{i,l}^2+f_{i,v}^2}\quad\forall i=1\dots \nrot.
\end{align}\label{eq:invAlloc}%
\end{subequations}%
If the system is overactuated, the vector $\vec{b}\in\nR{\nrot}$ can be used to find solutions in the nullspace of $\allocationMatrix$. For $\vec{b}=\bm{0}$ the norm of $\thrustComponents$ is minimized. Note that, due to the geometrical relation between $\thrustComponents$ and $\tprops$ in \cref{eq:extendedThrust}, the minimization of $\thrustComponents$ corresponds to a minimization of $\tprops$. We will refer to this solution as the \emph{minimum norm} or \emph{optimal allocation}, resulting in the optimal commands denoted by $\actInput^*=\matrix{\tiltangles^*{}\transpose & \tprops^*{}\transpose}\transpose$.


This procedure has the following properties:
\begin{enumerate*}[label=(\roman*)]
\item the norm of the resulting thrusts is minimized,\label{it:minnorm}
\item it is instantaneous,\label{it:instantaneous}
\item for $\vec{b}=\vec{0}$ the mapping $\actWrench\rightarrow(\tiltAngles^*,\tprops^*)$ is bijective, i.e.  for each $\actWrench$ there is a unique set of commands $(\tiltAngles^*,\tprops^*)$\label{it:bijective}
\end{enumerate*}.

While \ref{it:minnorm} and \ref{it:bijective} often provide advantages, \ref{it:instantaneous} can cause difficulties during fast motions. When $\actWrench$ changes rapidly, it can result in unfeasible fast changes of actuator commands. This can pose difficulties in the common separation of a high-level wrench generation controller and a low-level allocation.
In the following section, we will address this problem by introducing the allocation in the MPC formulation to impose cost and constraints on the actuator dynamics.

\section{Model Predictive Control Framework}

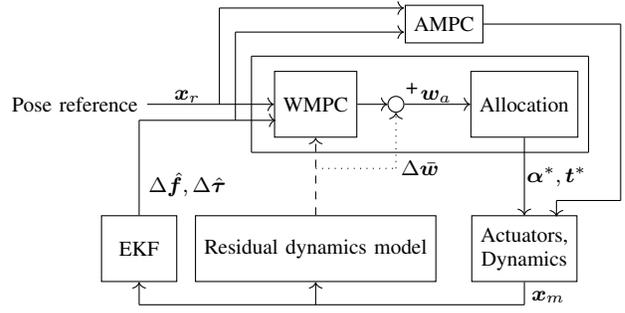
\begin{figure}[ht]
    \centering
    \begin{tikzpicture}[align=center, node distance=0.1cm, font=\footnotesize, scale=0.015cm]

    \node [style=smallBox] (5) at (0.5, 0) {WMPC};
    \node [style=smallBox] (6) at (7, 0) {Allocation};
    \node [style=smallBox] (8) at (7, -4.5) {Actuators,\\Dynamics};
    \node [style=smallBox] (9) at (0.5, -4.5) {Residual dynamics model};
    \node [style=smallBox] (10) at (-5, -4.5) {EKF};
    \node [style=whitebox] (11) at (-7, 0) {Pose reference};
    \node [style=circle] (12) at (3, 0) {};
    \node [style=none] (13) at (0.5, -2) {};
    \node [style=none] (14) at (3, -2) {};
    \node [style=none] (15) at (0.5, -6.25) {};
    \node [style=none] (16) at (-5, -6.25) {};
    \node [style=none] (17) at (-5, -0.5) {};
    \node [style=none] (18) at (7, -6.25) {};
    \node [style=none] (19) at (3.5, 0.5) {+};
    \node [style=none] (20) at (-0.75, -0.5) {};
    \node [style=none] (21) at (-3.5, -2.5) {$\Delta\hat{\bm{f}},\Delta\hat{\bm{\tau}}$};
    \node [style=none] (22) at (-3.5, 0.25) {$\bm{x}_r$};
    \node [style=none] (23) at (3.75, -2) {$\wrenchResidualModel$};
    \node [style=none] (24) at (8, -2.25) {$\bm{\alpha}^*, \bm{t}^*$};
    \node [style=none] (25) at (4.25, 0.25) {$\bm{w}_a$};
    \node [style=none] (26) at (0.5, -1) {};
    \node [style=none] (27) at (7.75, -6) {$\bm{x}_m$};
    \node [style=onelinebox] (28) at (4.5, 2.5) {AMPC};
    \node [style=none] (29) at (-1.5, 1.5) {};
    \node [style=none] (30) at (9, 1.5) {};
    \node [style=none] (31) at (9, -1.5) {};
    \node [style=none] (32) at (-1.5, -1.5) {};
	\node [style=none] (33) at (-2, 2.25) {};
    \node [style=none] (34) at (10, 2.5) {};
    \node [style=none] (35) at (10, -3) {};
    \node [style=none] (36) at (8, -3) {};
    \node [style=none] (37) at (-2, -0.5) {};
    \node [style=none] (40) at (8, -3.5) {};
    \node [style=none] (41) at (3.25, 3) {};
    \node [style=none] (42) at (-2.5, 0) {};
    \node [style=none] (43) at (3.25, 2.25) {};
    \node [style=none] (44) at (-2.5, 3) {};

    \draw [style=arrow] (11) to (5);
    \draw [style=arrow] (5) to (12);
    \draw [style=arrow] (12) to (6);
    \draw [style=dotted](13.center) to (14.center);
    \draw [style=arrow, dotted] (14.center) to (12);
    \draw (8) to (18.center);
    \draw (18.center) to (16.center);
    \draw (10) to (17.center);
    \draw [style=arrow] (15.center) to (9);
    \draw [style=arrow] (16.center) to (10);
    \draw [style=arrow] (6) to (8);
    \draw [style=arrow] (17.center) to (20.center);
    \draw [style=arrow, dashed] (9) to (5);
    \draw (29.center) to (30.center);
    \draw (30.center) to (31.center);
    \draw (31.center) to (32.center);
    \draw (32.center) to (29.center);
    \draw (37.center) to (33.center);
    \draw (28) to (34.center);
    \draw (34.center) to (35.center);
    \draw (35.center) to (36.center);
    \draw [style=arrow] (36.center) to (40.center);

    \draw [style=arrow] (33.center) to (43.center);
    \draw (42.center) to (44.center);
    \draw [style=arrow] (44.center) to (41.center);

\end{tikzpicture}\vspace{0.2cm}
    \caption{Control block diagram. Either the AMPC alone or the combination of WMPC and allocation can compute the actuator controls. The wrench residual $\wrenchResidualModel$ can be applied either directly in the WMPC formulation (\emph{In-MPC}, dashed) or added as a correcting feedforward-term (\emph{Post-MPC}, dotted). For AMPC, only residuals from the EKF can be applied.}\label{fig:wmpc_ampc_diagram}
\end{figure}

In this section, we present two different model-based controllers for the task of free flight trajectory tracking with an overactuated aerial vehicle.
Both methods are based on the same formalisms to model the system dynamics but differ in the level of detail to which the models are embedded and how residual wrenches are handled. Specifically, the first approach optimizes wrench commands and uses a default allocation, while the second approach operates on the actuator level. We refer to these two proposed methods as \ac{WMPC} and \ac{AMPC}, respectively.
Both controllers rely on residual wrench estimates to account for otherwise unmodeled disturbances. These estimates can either be provided by an online estimator (EKF) or by a Residual Dynamics Model (RDM).
\Cref{fig:wmpc_ampc_diagram} illustrates the two different methods in a control block diagram and \cref{tab:comparison} compares the similarities and differences between the two methods.
\begin{table*}[t!]
    \centering
    \begin{tabular}{c c c}
        \hline
         & \textbf{WMPC} & \textbf{AMPC}\\
        \hline 
        {Output} & Actuator wrench derivative $\actWrenchDot$ & Actuator commands derivative $\drotorSpeeds, \dtiltAngles$\\
        {Allocation} & Minimum norm allocation after MPC & Implicitly in MPC\\
        {Disturbance compensation} & Disturbance observer or model-based & Disturbance observer\\
        \hline 
    \end{tabular}\caption{Comparison of WMPC and AMPC.}
    \label{tab:comparison}
\end{table*}





\subsection{MPC formulation}
We first introduce a general formulation of the MPC problem.  To this end, we define the state vector as $\states\in\stateSpace\subseteq\nR{\nstates}$ and the input vector as $\inputs\in\inputSpace\subseteq\nR{\ninputs}$. 
We further assume that states and control inputs are constrained by the polytopes $\stateSpace$ and $\inputSpace$.
The system is subject to its dynamics $\dot\states=\vec{f}(\states,\inputs)$, discretized as $\states_{k+1}=\discreteDynamics(\states_k,\inputs_k)$.
We also define the stage cost $\stageCost(\states,\states_{\des})$ and terminal cost $\stageCost_N(\states_N,\states_{\des,N})$.
The discrete-time MPC problem is then formulated as the minimization of a cost function over a finite time horizon of $N$ steps:

\begin{subequations}
\begin{align}
\begin{split}
\min_{\inputs}\sum_{k=0}^{N-1}\biggl(\norm{\stageCost(\states_k,\states_{\des,k}&)}^2_{\mpcStateWeights}+\norm{\inputs_k}^2_{\mat{R}}\biggr)\\
&\quad+\norm{\stageCost_N(\states_N,\states_{\des,N})}^2_{\mpcStateWeights_N}
\end{split}\label{eq:mpc_optimization}\\
\begin{split}
\text{subject to}\quad&\states_k\in\stateSpace, \inputs_k\in\inputSpace\\
&\states_{k+1}=\discreteDynamics(\states_k,\inputs_k)\\
&\states_0=\states(t).
\end{split}\label{eq:mpc_subject}%
\end{align}\label{eq:mpc}%
\end{subequations}%
The matrices $\mpcStateWeights,\ \mpcStateWeights_N\in\nR{\nstates\times \nstates}$ and $\mpcInputWeights\in\nR{\ninputs\times \ninputs}$ represent the state, terminal state, and input cost matrices, respectively.
For the remainder of this section we present the details and differences of the two approaches and how \cref{eq:mpc} is adapted accordingly.
\subsection{State and input vectors for different model formulations}
For the two formulations of the \ac{WMPC} and \ac{AMPC} we use different definitions of state and input vectors.
\subsubsection{Wrench-MPC}
In the case of \ac{WMPC} we define the state vector to comprise both the \ac{OMAV} and the wrench states. The modeling of the state dynamics is equal to the rigid body dynamics in \cref{eq:systemDynamicsShort}.
\begin{subequations}
\begin{align}
\states_W&=\matrix{\actWrench\transpose & \pos\transpose & \vel\transpose & \qB\transpose & \angVel\transpose}\transpose\in\nR{19} \\
\inputs_W&=\actWrenchDot\in\nR{6}.
\end{align}
\end{subequations}
\subsubsection{Actuator-MPC}
While the system dynamics in \ac{AMPC} are described equivalently to \ac{WMPC}, we include the actuator commands $\actInput=\matrix{\tiltangles\transpose & \tprops\transpose}\transpose$ in the state as well as the allocation \cref{eq:allocation} in the MPC system dynamics.
\begin{subequations}
\begin{align}
\states_A&=\matrix{\tiltangles\transpose & \tprops\transpose & \pos\transpose & \vel\transpose & \qB\transpose & \angVel\transpose}\transpose\in\nR{31} \\
\inputs_A&=\matrix{\tiltanglesdot\transpose & \tpropsdot\transpose}\transpose = \actInputDot \in\nR{18}.
\end{align}
\end{subequations}
By including the actuator commands in the model, the MPC does not rely on the allocation procedure \cref{eq:invAlloc}, thus allowing a larger exploration space of possible solutions. Additionally, we can impose constraints on actuator velocities and ensure continuity of the commands.

\subsection{Cost vector}
The cost vector $\stageCost(\states,\states_{\des})$ is designed slightly different for the two different approaches of \ac{WMPC} and \ac{AMPC}. Both have in common that a reference trajectory is to be tracked, given by the time dependent variables $\pos_\des,\vel_\des,\qB_\des,\angVel_\des$.
\subsubsection{Wrench-MPC}
We employ a common definition of tracking errors to write the cost vector:
\begin{align}
\stageCost(\states_k,\states_{\desk})=\matrix{\pos_k-\pos_{\desk} \\ \vel_k-\vel_{\desk} \\ \qErrk \\ \angVel_k-\RB \transpose \RBdes \angVel_{\desk}},
\end{align}
with $\qErr\in\nR{3}$ as the vector that describes the rotation error between $\qB$ and $\qB_\des$, such that
\begin{equation}
\qB_\des=\qB\otimes\Delta\qB,\quad \Delta\qB\coloneqq \matrix{1 \\ \qErr}.
\end{equation}
\subsubsection{Actuator-MPC}
The actuator-based MPC further includes the tilt angles and propeller thrusts as state variables:
\begin{align}
    \stageCost(\states_k,\inputs_k)=\matrix{\pos_k-\pos_{\desk} \\ \vel_k-\vel_{\desk} \\ \qErrk \\ \angVel_k-\RB \transpose \RBdes \angVel_{\desk} \\ \tiltAngles_k-\tiltAngles_{\desk} \\ \tprops_k-\tprops_{\desk}}.
\end{align}
Generally, we aim to allow the actuator dynamics to evolve as freely as possible. There are different possibilities to penalize these states.
Not penalizing the actuator states would allow the largest exploration freedom but would could also lead to long optimization times of the MPC.
Therefore, we chose to penalize the deviation of the actuator commands from the minimum norm commands $(\tiltAngles^*,\tprops^*)$:
\begin{subequations}
    \begin{align}
        \tiltAngles_{\desk} &= \tiltAngles^*_t\\
        \tprops_{\desk} &= \tprops^*_t.
    \end{align}
\end{subequations}
Tuning of the weight matrix $\mpcStateWeights$ then allows to give more or less range in deviating from the optimal allocation.
The minimum norm commands $(\tiltAngles^*_t,\tprops^*_t)$ are obtained assuming static hover, i.e., only exerting the force required to hold the platform weight in the current attitude.

\subsection{Constraints}
In order to obtain smooth control inputs and to account for unmodeled dynamic effects, we can employ hard constraints on any of the states and inputs.
\subsubsection{\ac{WMPC}}
In the case of WMPC, we constrain the total actuator wrench $\actWrench$ and its derivative $\actWrenchDot$. This allows us to ensure that both the total wrench and the wrench rate remain in feasible bounds. Note that employing constraints on the wrench rate implicitly constrains the actuator rates due to the relation in \cref{eq:allocation}.
\begin{subequations}
    \begin{align}
    -\actWrenchDotConstraint \leq \actWrenchDot \leq \actWrenchDotConstraint\\
    -\actWrenchConstraint \leq \actWrench \leq \actWrenchConstraint
    \end{align}
\end{subequations}

\subsubsection{\ac{AMPC}}
One major advantage of AMPC is the possibility to constrain actuator commands directly. Therefore, we employ hard constraints on the thrusts, the thrust rates, and the tilt angle rates:
\begin{subequations}
    \begin{align}
        \tpropsMin \leq {}&\tprops \leq \tpropsMax\\
        -\tpropsdot_{max} \leq {}&\tpropsdot \leq \tpropsdot_{max}\\
        -\tiltanglesdot_{max} \leq {}&\tiltanglesdot \leq \tiltanglesdot_{max}
    \end{align}
\end{subequations}

\subsection{Optimal problem result}
Each MPC iteration of solving \cref{eq:mpc} returns a sequence of optimal inputs $U=\left[\inputs_0^*,\dots,\inputs_N^*\right]$ and associated states $X=\left[\states_0^*,\dots,\states_N^*\right]$. We use this state sequence to extract the optimal inputs and apply them as control inputs to the system --- either $\actWrench$ in the case of WMPC or $\actInput$ in the case of AMPC.

\section{Compensation for disturbances}
We identify model mismatches as a main cause that leads to non-optimal tracking of reference trajectories. Therefore, we introduce two methods to tackle this challenge: \begin{enumerate*}[label=(\roman*)]
    \item an EKF-based disturbance observer, and
    \item a linear model that predicts disturbance wrenches based on experimental data.
\end{enumerate*}
Both methods intend to predict the residual wrench $\wrenchResidual=\matrix{\forceResidual\transpose & \torqueResidual\transpose}\transpose$ for each flight configuration.

\subsection{Disturbance observer}
We employ an \ac{EKF} to estimate the disturbance force and torque in real time. 
We assume that most disturbances originate from internal model errors, e.g., from interfering air flows, inaccurate rotor-speed/thrust mapping, or misaligned tilt arms. Furthermore, we assume that these internal errors are independent of the platform yaw angle $\psi$.
Therefore, we estimate the disturbance force in the local frame $\frame{L}$, which is obtained by a pure yaw rotation of the platform yaw angle from the world frame, i.e., $\rotMat_L=\rotMat_z(\psi)$.
We use the following state vector $\stateEst\in\nR{19}$, inputs $\ekfInputs\in\nR{6}$, and measurements $\ekfMeasurements\in\nR{7}$:
\begin{align}
\stateEst=\matrix{\posEst\\ \velEst \\ \qBEst \\ \angVelEst \\ \distForceEst \\ \distTorqueEst},\quad\ekfInputs=\matrix{\actForce\\ \actTorque},\quad\ekfMeasurements=\matrix{\posMeas\\ \qBMeas}.
\end{align}
The formulation of linear and rotational dynamics is equal to \cref{eq:systemDynamicsShort}, while the evolution of the disturbance force and torque is assumed to be constant, i.e.
\begin{subequations}
\begin{align}
\ddt\distForceEst&=\processNoise_{\distForceEst}\\
\ddt\distTorqueEst&=\processNoise_{\distTorqueEst},
\end{align}
\end{subequations}
where $\processNoise_{\distForceEst}\sim\mathcal{N}(\zeros,\mat{\Sigma}_f),\ \processNoise_{\distTorqueEst}\sim\mathcal{N}(\zeros,\mat{\Sigma}_\tau)$ represent the process noise, respectively.
We obtain the disturbance force in the body frame by the following rotation:
\begin{align}
    \forceResidualEst&=\RB\transpose\rotMat_L \distForceEst.
\end{align}
The force and torque disturbance estimates are directly employed in the dynamic model of the MPC formulation, specifically in \cref{eq:systemDynamicsShort}.

\subsection{Residual Dynamics Model (RDM)}
The above introduced method of estimating disturbances online comes with the downside of being time-dependent and, therefore, can introduce time delays. Therefore, we now present another method which relies on estimating the internal disturbances based on a model which is trained by experimental data.
To this end, we approximate the true residual wrench $\wrenchResidual$ with a linear model $\wrenchResidualModel=f(\features)$, with $\features$ as a feature vector.
We follow a similar approach as in \cite{Aswani2012} with the difference of learning the parametric uncertainties offline rather than in-flight. This has the advantage that the parameters are not estimated online which could lead to unpredictable and inconsistent flight behavior.

\subsubsection{Model definition}
We use a feature vector $\features\in\nR{\nfeatures}$, to create a linear affine relationship between a set of $\nfeatures$ features and the residual wrench:
\begin{align}
\wrenchResidualModel(\features)=\modelMatrix\features,
\end{align}
where $\modelMatrix\in\nR{6\times \nfeatures}$ is a matrix that maps from features to wrench residuals.
The choice of a simple linear model allows us to employ it in the MPC framework while maintaining a low computational complexity.



\subsubsection{Feature selection}
Selecting an appropriate set of features is important to capture relationships between available data and perceived dynamic residuals while keeping the mathematical complexity low.
We will present our selection of features in the experimental section \ref{sec:wmpcmodel}.



\subsubsection{Training}
In this section we describe the process of finding the model matrix $\modelMatrix$ 
for an optimal performance when employing the learned model in the control loop.


Given a dataset of $\nsamples$ experimentally recorded residual wrenches $\wrenchResidual_{m,i}$ and features $\features_i,\ i\in{1,\dots,\nsamples}$ we want to find $\modelMatrix$ s.t.
\begin{align}
    \modelMatrix&=\argmin_{\modelMatrix}\sum_{i=1}^{\nsamples}\trainError{}_{,i}\\
    \trainError{}_{,i}&=\norm{\modelMatrix\features{}_i-\wrenchResidualMeasured{}_{,i}}.
\end{align}
\subsubsection{Computation of dynamics residuals}
We use linear acceleration and angular velocity data obtained from an onboard \ac{IMU}, $\imuLinAccMeas$ and $\imuAngAccMeas$, respectively, to compute the residuals from recorded training datasets. The recorded angular velocity $\imuAngVel$ is differentiated numerically to obtain the angular acceleration $\imuAngAcc$.

Furthermore, we employ MSF \cite{Lynen2013} to correct the linear acceleration measurements for the IMU bias.

We can then compute the wrench residuals from the measured accelerations as
\begin{align}
\wrenchResidualMeasured=\matrix{\forceResidualMeasured \\ \torqueResidualMeasured}=\matrix{\mass\imuLinAccMeas-\actForce \\ \inertia\imuAngAcc-\actTorque}
\end{align}

In order to compute the model parameters we use ridge regression to minimize the training error. We train each row of the model matrix individually. Let us define the feature matrix for $\nsamples$ samples as $\featureMatrix\in\nR{\nsamples\times\nfeatures}$ and the vector of measured residuals for the $i$-th component of the wrench $\vec{y}_i$:
\begin{align}
\featureMatrix=\matrix{\features_0 & 1\\ \features_1 & 1 \\ \vdots \\ \features_{\nsamples-1} & 1},\quad \vec{y}_i=\matrix{\wrenchResidual_{i,0}\\ \wrenchResidual_{i,1}\\ \vdots\\ \wrenchResidual_{i,\nsamples-1}},\quad\modelMatrix=\matrix{\vec{c}_0 \\\vec{c}_1 \\ \vdots \\ \vec{c}_5},
\end{align}
where $\vec{c}_i\in\nR{1\times \nfeatures}$ represents the $i$-th row of the model matrix. 
We can then find the model coefficients for each wrench component individually:
\begin{align}
\vec{c}_i=\argmin_{\vec{c}_i} \norm{\vec{y}_i-\featureMatrix\vec{c}_i\transpose}+\regParam\norm{\vec{c}_i},\quad i\in\{0,\dots,5\}.\label{eq:ridgeRegression}
\end{align}
Note that \cref{eq:ridgeRegression} is a ridge regression with $\regParam$ as the regularization parameter that helps avoid overfitting to the training data.



\subsubsection{Application of the residuals in the control loop}
Generally, we only employ the residual model in the WMPC formulation only and not in AMPC. 
This is because the allocation nullspace exploitation in AMPC can result in various different actuator commands $\actInput$ for the same states, leading to different wrench residuals as a consequence and making a parametric model infeasible. 

Within WMPC, we investigate two different methods of applying the residual model in the control framework. 
\paragraph{In-MPC} In this approach, the model is implemented in the state dynamics of the MPC formulation. This allows the controller to respect the residual dynamics while computing an optimal input trajectory.
\paragraph{Post-MPC} In this approach, the MPC is agnostic of any model inaccuracies. Instead, the resulting optimal wrench commands $\actWrench^*$ are corrected after the MPC optimization.


\subsection{Discussion}
The first approach of employing an EKF-based disturbance observer has the advantage of being simple to implement while being able to adapt to most disturbances. On the other hand, it introduces a time delay into the system as it requires sensor measurements to adapt its estimates.

The model-based approach can be instantly applied in the controller framework, not adding any time delays. However, it requires a rich dataset upon which the model parameters can be fit. Additionally, selecting the correct features and a reasonable regularization parameter is not straight forward. Because of its linear formulation it can also only cover a limited area around a specific operating point.

\section{Experimental validation}

\subsection{Implementation}

\subsubsection{Flying platform}
We perform all experiments on our custom built \ac{OMAV}. This platform is designed with 6 arms equally spaced around its body center. Each arm can be tilted individually by a Dynamixel XL430-W250-T servo. At the end of each arm a double rotor group, containing two KDE2315XF-885 motors with counter rotating 9x\SI{4.5}{in} propellers is mounted.  The counter rotation of each rotor group minimizes the net torque of each arm, but the exact influence of the airflow interferences is unknown.
The entire system is powered by a single \SI{7000}{\milli\amperehour} battery. Fully prepared for a flight its mass is $\SI{4.36}{\kilo\gram}$.

\subsubsection{Software}
The entire controller is implemented in ROS on an Intel NUC that is mounted on the platform.  Reference trajectories are transmitted via WiFi from an offboard computer.
The onboard computer runs the MPC solver and publishes either wrench or actuator commands (according to WMPC or AMPC), which are forwarded to a Pixhawk flight controller. For WMPC, the flight controller computes the optimal actuator commands and sends them to the actuators, while for AMPC it solely passes the commands through to the actuators.

The MPC optimizer is implemented using the ACADO framework. The system dynamics are discretized through direct multiple shooting and solved through an Implicit Runge Kutta method (Gauss-Legendre integrator of order 6). We use qpOASES as the QP solver.

\subsection{Model for WMPC}\label{sec:wmpcmodel}
We have tested different feature sets for the model.  For the experiments we used a set made up from the commanded wrench and the roll and pitch angle, encoded by the 3rd row of the rotation matrix $\RB$, resulting in 9 features:
\begin{equation}
\features=\matrix{\actWrench\transpose & -\sin(\theta) & \cos(\theta)\sin(\phi) & \cos(\theta)\cos(\phi)}\transpose.
\end{equation}
It turned out that the regularization parameter $\regParam$ is an essential tuning parameter for the closed loop stability.
Low values of $\regParam$ lead to a better model fit to the training data,  but also to high residual predictions, resulting in strong countersteering and increasing instabilities. We therefore converged to choosing a high value of $\regParam=\SI{1e5}{}$.

\begin{figure}[t]
    \includegraphics[width=\columnwidth]{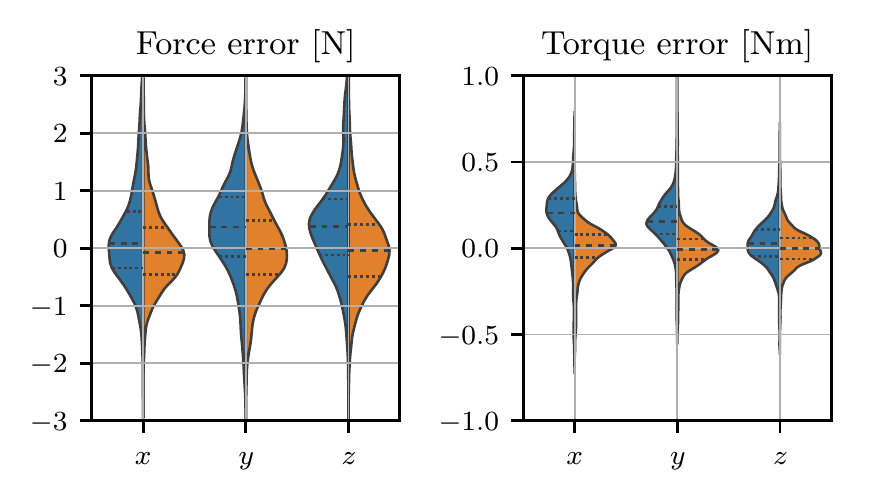}\caption{Model fit errors for residual forces and torques. The blue violinplots represent the raw residual data $\wrenchResidualMeasured{}_{,i}$ and the orange violinplots represent the residuals after fitting $\trainError{}_{,i}$.}\label{fig:modelErrors}
\end{figure}

We recorded a training dataset which consists of two trajectories: the first one containing pure pitching and rolling motions of with a duration of \SI{167}{\second} and the second one tracking out a horizontal square, lasting \SI{178}{\second}.
This dataset therefore contains both angular as well as linear acceleration data.
The results of the model fit are presented in \cref{fig:modelErrors} and \cref{tab:modelErrors}.
Both the table and the figure show the RMSE of the training data, i.e. of $\trainError$. It can be seen that the model is able to compensate especially static offsets in both force and torque.

\begin{table}[t!]
\centering
\begin{tabular}{c c c}
\hline 
 & RMSE force [\si{\newton}] & RMSE torque [\si{\newton\meter}] \\ 
\hline 
Raw & $1.444\pm 0.867$  & $0.347\pm 0.147$ \\
Model ($\regParam=\SI{1e5}{}$) & $1.144\pm 0.669$ & $0.177\pm 0.115$ \\
Exact Model ($\regParam=0$) & $0.959\pm 0.517$ & $0.087\pm 0.057$ \\
\hline
\end{tabular}\caption{Force and torque RMSE and std. deviations of raw data and after model fit.}
\label{tab:modelErrors}
\end{table}

\subsection{Experiments}
The presented building blocks of WMPC/AMPC and disturbance observer/parametric model are combined in different ways to evaluate their performances in real-world experiments. 

Specifically, we aim to analyze the following characteristics: \begin{enumerate*}[label=(\roman*)]
    \item Position and attitude tracking performance, and
    \item velocity of actuator commands as well as nullspace exploitation of the AMPC.
\end{enumerate*}

\begin{figure}[t]
        \centering
        \includegraphics{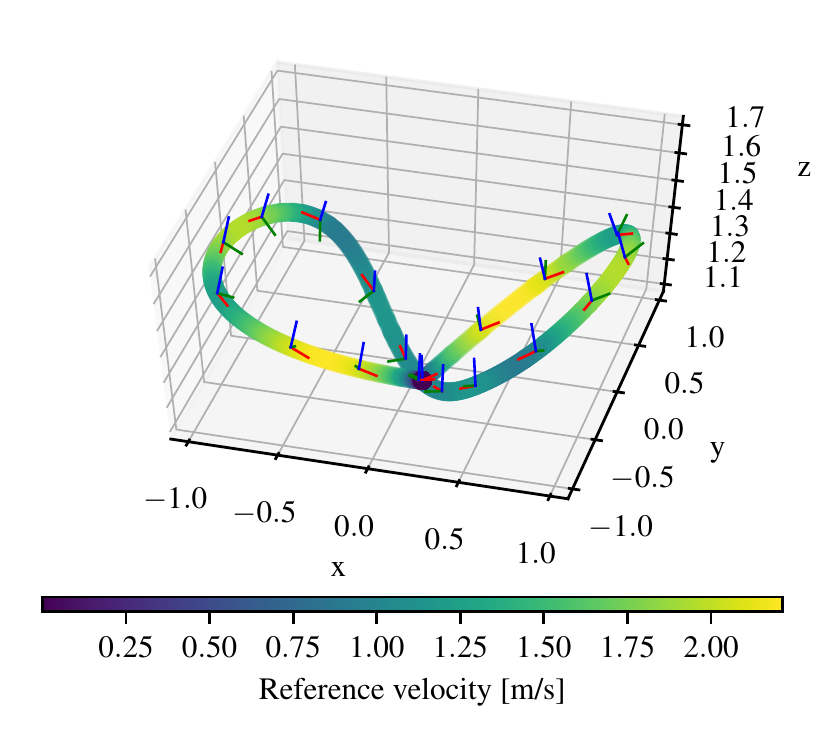}
        \caption{Bent lemniscate trajectory. Note that this trajectory also involves pitch angles up to \SI{30}{\degree}.}\label{fig:lemniscate}
\end{figure}

We evaluate the controllers by comparing their capabilities to track given 6-\ac{DOF} trajectories in free space with an \ac{OMAV}.
We use four different trajectories and different velocities to evaluate the tracking performance: \begin{enumerate*}[label=(\roman*)]
\item square trajectory: horizontal square with \SI{1}{\meter} leg length and reference velocities up to \SI{3}{\meter\per\second},
\item attitude trajectory: pure pitching and rolling up to \SI{45}{\degree} while hovering at a fixed position with a duration of \SI{27}{\second},
\item lemniscate trajectory: combined position and attitude trajectory tracking a bent lemniscate with a two possible speeds (slow: \SI{15}{\second} duration, up to \SI{0.9}{\meter\per\second} and fast: \SI{5.5}{\second}, up to \SI{2.9}{\meter\per\second}), as presented in \cref{fig:lemniscate}, and
\item horizontal step responses along the $x$-axis with \SI{1}{\meter} length.
\end{enumerate*}
We compute the attitude errors as euler angles of the actual attitude w.r.t. the \emph{reference} attitude. That way, we avoid large or distorted angle errors at large roll/pitch angles. Accordingly, the attitude RMSE is the RMSE of the error euler angles.

\Cref{tab:settings} presents the values of the most important tuning parameters. Note that the horizon length for AMPC is shorter than for WMPC in order to keep the computational complexity low. In both cases we use a time discretization of \SI{50}{\milli\second}, resulting in time horizons of \SI{1}{\second} for WMPC and \SI{0.5}{\second} for AMPC, respectively.

\begin{table}[t!]
    \centering
    \begin{tabular}{c c c}
    \hline 
    Parameter & WMPC & AMPC \\ 
    \hline 
    $N$ & 20 & 10 \\
    $\Delta t$ & \SI{0.05}{\second} & \SI{0.05}{\second}\\
    $\actForceConstraint$ & \SI{20}{\newton} & \SI{20}{\newton} \\
    $\actTorqueConstraint$ & \SI{20}{\newton\meter} & \SI{20}{\newton\meter} \\
    $\tiltanglesdot_{max}$ & n/a & \SI{10}{\radian\per\second}\\
    $\tprops_{max}$ & n/a & \SI{16}{\newton}\\
    $\tprops_{min}$ & n/a & \SI{0.1}{\newton}\\
    $\tpropsdot_{max}$ & n/a & \SI{29}{\newton\per\second}\\
    \hline 
    \end{tabular}\caption{Control parameters for WMPC and AMPC.}
    \label{tab:settings}
\end{table}

\subsection{WMPC}
\begin{figure}[h]
\includegraphics[width=\columnwidth]{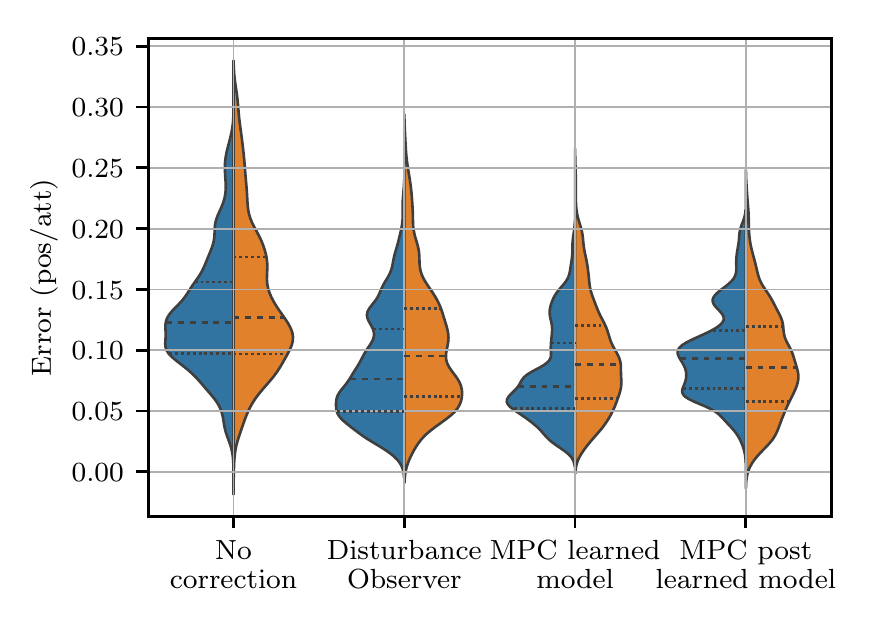}\caption{Tracking errors of the WMPC during the attitude trajectory for different controller configurations. The violins show the position RMSE on the left (in [\si{\meter}]) and the attitude RMSE on the right (in [\si{\radian}]), respectively.}\label{fig:wmpc_te_violin}
\end{figure}

We evaluate the different variations of WMPC (i.e., no correction for disturbances, residual model in- or post-MPC, or online disturbance observer) by tracking different trajectories. Specifically, we focus on the influence of the disturbances on the tracking performance, and how well these can be compensated by the linear model approach.
\Cref{tab:trackingWMPC} gives an overview of the RMSE of each experiment. It shows that in most cases the Post-MPC variant outperforms all other configurations by a small margin. This is also highlighted in \cref{fig:wmpc_te_violin}, where we show the pose tracking errors only for the attitude trajectory in different configurations.

\begin{table}[t!]
\centering
\begin{tabular}{c|c|c|c|c}
\hline 
Position err. [\si{\meter}] & N/c & In-MPC & Post-MPC & D/o \\ 
\hline 
Squares & 0.198 & 0.172 & \textbf{0.150} & - \\
Attitude trajectory & 0.139 & \textbf{0.088} & 0.104 & 0.095 \\
Lemniscate & 0.137 & 0.091 & \textbf{0.085} & 0.100 \\
Lemniscate fast & 0.146 & 0.109 & \textbf{0.108} & - \\
\hline 
Attitude err. [\si{\radian}] & N/c & In-MPC & Post-MPC & D/o \\ 
\hline 
Squares & 0.210 & 0.174 & \textbf{0.167} & - \\
Attitude trajectory & 0.152 & 0.104 & \textbf{0.100} & 0.116\\
Lemniscate & 0.167 & 0.106 & \textbf{0.105} & 0.123\\
Lemniscate fast & 0.215 & \textbf{0.140} & 0.156 & - \\
\hline 
\end{tabular}\caption{Trajectory tracking RMSE of the WMPC approach for different controller configurations and trajectories. N/c = ``No correction'' (i.e., no disturbance compensation).}
\label{tab:trackingWMPC}
\end{table}

\subsection{AMPC}
\begin{figure}[ht]
    \includegraphics{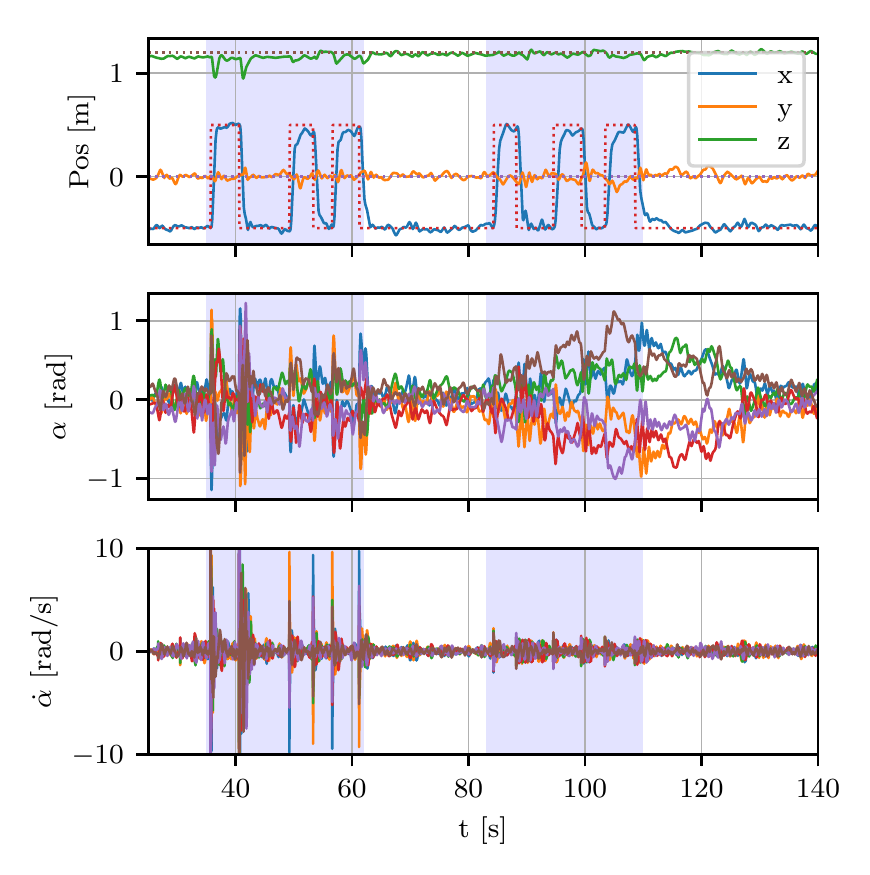}
    \caption{Position tracking, tilt angle commands and tilt angle speeds for lateral position steps. The actuator weights are high in the first highlighted period and low in the second highlighted period.}\label{fig:tiltAnglesAMPC}
\end{figure}

For AMPC, we analyze the influence of the controller tuning (i.e., the actuator input weights) on the pose tracking performance and on the exploitation of the allocation nullspace. To this end, we perform two sets of experiments. In both sets, we first track the reference trajectory with high weights $\weightAlpha$ and then with low weights.
\begin{enumerate*}
    \item We first apply this procedure on the square trajectory. The numerical results in \cref{tab:trackingAMPC} show that the tracking accuracy is higher for high actuator weights.
    \item We then also use this procedure on tracking horizontal position reference steps of \SI{1}{\meter}. Horizontal steps are particularly challenging as they require a sudden thrust direction change that can lead to infeasibly high tilt angle speeds. \Cref{fig:tiltAnglesAMPC} presents the results with a focus on the tilt angle commands. It shows that for low actuator weights, the tilt angle commands change rapidly and exhibit infeasibly high velocities. However, for higher weights, the tilt angle speeds are significantly lower while the position tracking is only slightly affected, resulting in \SI{0.328}{\meter} in the first and \SI{0.351}{\meter} in the latter case, respectively. Furthermore, note the tilt angle drift in the second period. As the weights $\weightAlpha$ are lowered, the allocation nullspace is explored more freely, neglecting the objective to achieve maximum power efficiency.
\end{enumerate*}

\begin{table}[t!]
\centering
\begin{tabular}{c|c|c|c|c|c}
\hline 
 & $\weightThrust$ & $\weightAlpha$ & $\weightAlphaDot$ & Pos. err. [\si{\meter}] & Att. err. [\si{\radian}]\\ 
\hline 
High alpha cost & 1.0 & 10 & 10 & 0.111 & 0.212 \\
Low alpha cost & 0.1 & 0.1 & 10 & 0.136 & 0.279 \\
\hline 
\end{tabular}\caption{Trajectory tracking RMSE for different controllers and trajectories. For both AMPC tunings, the square trajectory was tracked three times, resulting in evaluation times of \SI{30}{\second}.}
\label{tab:trackingAMPC}
\end{table}


\subsection{Comparison}
\begin{figure}[ht]
    \includegraphics{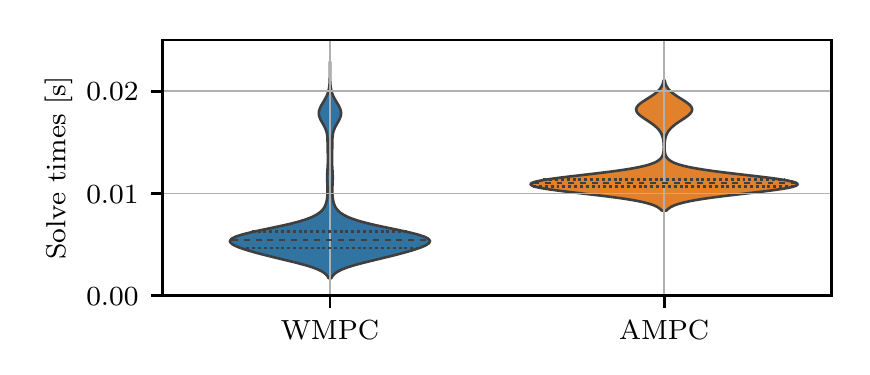}
    \caption{Comparison of MPC solver times for WMPC and AMPC.}\label{fig:solveTimes}
\end{figure}

Both presented methods have their respective advantages and disadvantages. 
\begin{itemize}
    \item Implementation and user-friendliness: Generally, both WMPC and AMPC are similarly complex in their implementation on a flying platform. However, as the input dimensionality of AMPC is considerably larger, tuning weights and constraints can be more tedious as compared to WMPC. Additionally, as AMPC can exploit the entire allocation nullspace of an overactuated vehicle, the flight behavior can be inconsistent and produce non-repeatable results.
    \item Tracking accuracy: Using the Post-MPC formulation in WMPC we have found the highest tracking accuracy. This is due to the no-delay advantage of a learnt model (as opposed to an online filter) and the relatively accurate allocation model at the optimal solution. While AMPC in theory should perform better, we suspect that the combination of longer computation times, the delay produced by the EKF, and the allocation model being inaccurate far away from the optimal solution lead to higher tracking errors.
    \item Power efficiency: As WMPC uses a maximum-power-efficiency allocation, it is more efficient than AMPC, which also produces suboptimal control inputs for the purpose of complying with actuator constraints and the input weights.
    \item Applications: In most applications, WMPC provides a sufficient performance. However, AMPC can be of interest in the case of highly aggressive maneuvers in which actuator constraints need to be considered or in which the optimal allocation solution alone does not produce satisfying results, requiring the exploitation of the allocation nullspace.
    \item Computational complexity: Due to the larger input and state space of AMPC, it takes considerably longer computation times. \Cref{fig:solveTimes} shows that AMPC exceeds the desired computation time of \SI{10}{\milli\second}, especially during periods in which constraints are active. As an example, the second period of \cref{fig:tiltAnglesAMPC} leads to the long solver times of nearly \SI{20}{\milli\second}.
\end{itemize}

\section{Conclusion}
We have presented a model predictive control framework for fully actuated or overactuated \acp{MAV}. Within this framework, we have employed two \acp{MPC} that optimize different control inputs and that use different approaches to cope with disturbances that arise from unknown internal effects.

The first one, WMPC, optimizes actuator wrenches which are thereafter translated by an optimal allocation into actuator commands. WMPC can consider disturbances either inside the model formulation (In-MPC) or as a a posteriori correction of the optimal wrench commands (Post-MPC). The disturbances can either be estimated by an EKF or by an approximation through a linear model that is trained on experimental data.

On the other hand, AMPC optimizes actuator commands and relies on an EKF as a disturbance estimator. Due to its knowledge of the actuator allocation, it can exploit the allocation nullspace and direct constraints on actuator commands.

Finally, we have conducted experiments to show the performances of the two controllers and their respective up- and downsides. While the AMPC approach in theory models the system more accurately, it suffers from the higher complexity (both in tuning, repeatability, and in computation). Therefore, the WMPC appraoch remains the preferred method for most use cases.

\bibliographystyle{IEEEtran}
\bibliography{./bibCustom}

\newpage



\end{document}